\documentclass{article}

\usepackage{microtype}
\usepackage{graphicx}
\usepackage{subfigure}
\usepackage{booktabs}

\usepackage{hyperref}

\usepackage[accepted]{icml2026}

\AtBeginDocument{%
  \fancyhf{}%
  \fancyhead[LO,LE]{\fontsize{8}{8}\selectfont
    Published as a workshop paper at SCALE - ICML 2026}%
  \fancyfoot[C]{\fontsize{8}{8}\selectfont\thepage}%
}

\usepackage{amsmath}
\usepackage{amssymb}
\usepackage{mathtools}
\usepackage{amsthm}

\usepackage{algorithm}
\usepackage{algorithmic}

\usepackage[capitalize,noabbrev]{cleveref}

\usepackage{enumitem}
\usepackage{xspace}

\usepackage{tikz}
\usetikzlibrary{arrows.meta,positioning,calc}

\theoremstyle{plain}

\theoremstyle{definition}

\theoremstyle{remark}

\newcommand{\TTR}{\ensuremath{\mathrm{TTR}}\xspace}
\newcommand{\rhokey}{\ensuremath{\rho_{\mathrm{key}}}\xspace}
\newcommand{\Lstar}{\ensuremath{L^{*}}\xspace}
\newcommand{\Mllm}{\ensuremath{M_{\mathrm{LLM}}}\xspace}
\newcommand{\Mcomp}{\ensuremath{M_{\mathrm{comp}}^{(p)}}\xspace}
\newcommand{\Mkv}{\ensuremath{M_{\mathrm{KV}}^{(p)}}\xspace}
\newcommand{\LLMTwo}{LLMLingua\nobreakdash-2\xspace}

\icmltitlerunning{Tri-Metric Routing Framework for Long-Context RAG on Commodity GPUs}

\begin{document}

\twocolumn[
\icmltitle{Beyond Static RAG: An Adaptive, Tri-Metric Routing Framework\\
           for Efficient Long-Context Inference on Commodity GPUs}

\icmlsetsymbol{equal}{*}

\begin{icmlauthorlist}
\icmlauthor{Saipraveen Vabbilisetty}{inst1}
\icmlauthor{Ajay Kumar Boddepalli}{inst1}
\icmlauthor{Deep Narayan Mishra}{inst1}
\icmlauthor{Shashank Kapadia}{inst1}
\icmlauthor{Haoan Wang}{inst1}
\icmlauthor{Anupriya Sharma}{inst1}
\end{icmlauthorlist}

\icmlaffiliation{inst1}{Walmart Inc., Sunnyvale, California, United States of America}

\icmlcorrespondingauthor{Saipraveen Vabbilisetty}{Saipraveen.Vabbilisetty0@walmart.com}

\icmlkeywords{retrieval-augmented generation, agentic AI systems, agentic memory,
  hardware-aware co-design, context compression, adaptive execution, commodity GPU}

\vskip 0.3in
]

\printAffiliationsAndNotice{}

\begin{abstract}
Deploying retrieval-augmented generation (RAG) pipelines on commodity
accelerators such as the NVIDIA T4 (16\,GB VRAM) exposes a critical
failure mode we term the \textbf{Compression Paradox}: neural prompt
compressors introduce key-value (KV) cache contention and latency
overhead that frequently \emph{exceed} the generation time saved, while
bypassing the compressor causes out-of-memory (OOM) crashes on long
inputs.
We identify two mechanistically distinct failure modes arising from
co-deploying a vLLM\nobreakdash-served LLM~\cite{kwon2023pagedattention}
and a PyTorch neural compressor on constrained hardware, and propose
the \textbf{Tri-Metric Router}---a deterministic, training-free dispatch
policy that selects among Raw, Neural (LLMLingua\nobreakdash-2~\cite{pan2024llmlingua2}),
and Lexical (BM25~\cite{robertson2009bm25}) pipelines using three
CPU-bound signals: spatial complexity $L$, syntactic density
$\rho_{\mathrm{key}}$, and Type-Token Ratio (\TTR).
The adaptation signal is \emph{hardware-physical}---VRAM headroom and a
latency crossover---rather than semantic, an underexplored axis of
efficient inference on commodity accelerators.
Routing thresholds are derived from a profiling sweep on LongBench
\texttt{qasper}~\cite{bai2024longbench}, isolating a latency crossover
at $\Lstar \approx 4{,}332$ words on the T4; the contribution is the
calibration \emph{methodology}, not the hardware-specific scalar.
The router achieves \textbf{0\% OOM failures} and
\textbf{88.5\,$\pm$\,4.4\% oracle alignment on out-of-distribution holdouts}
with a statistically stable \textbf{49.3\% Combined F1}---a 5.2-point
gain over always-on lexical compression at zero additional VRAM or
training cost.
\end{abstract}

\section{Introduction}
\label{sec:intro}

The NVIDIA T4 is the dominant commodity accelerator across cloud spot
markets, academic clusters, and cost-sensitive enterprise deployments.
Retrieval-augmented generation~\cite{lewis2020rag} and
passage-fusion architectures such as Fusion-in-Decoder~\cite{izacard2021fid}
have become standard paradigms for knowledge-intensive NLP, concatenating
multiple retrieved passages into long contexts that make efficient
inference on constrained hardware a pressing deployment challenge.
Its 16\,GB GDDR6 ceiling imposes inference economics that differ
qualitatively from the A100/H100 baselines on which state-of-the-art
context compression is typically evaluated~\cite{pan2024llmlingua2}.
We address this with a deterministic, training-free dispatch
policy---analytic and auxiliary-model-free at dispatch time, with no
learned components and no auxiliary model training---whose novelty
lies not in a new model but in applying a \emph{hardware-physical}
adaptation signal to a problem previously treated as purely semantic.

\paragraph{The Compression Paradox}
Lla\-ma-3-8B-In\-struct-AWQ~\cite{lin2024awq} served via
vLLM~\cite{kwon2023pagedattention} leaves ${\sim}7$\,GB for auxiliary
models; \LLMTwo consumes 2.5\,GB of this statically.
Beyond $\approx$4{,}300 words two failure modes converge:
(i)~latency inversion, where encoder overhead exceeds generation time
saved; and (ii)~\texttt{Cache\-Engine} block pool exhaustion
(Always-Raw) and cross-process PyTorch CUDA OOM (Always-Neural),
as vLLM's rigid 55\% reservation starves the encoder's allocator.
We call this the \textbf{Compression Paradox}: invoking a compressor
renders inference \emph{slower and less reliable} than raw passthrough.

\paragraph{Limitations of prior approaches}
Adaptive-RAG systems~\cite{asai2024selfrag,jeong2024adaptiverag,
jiang2023flare,li2024selfroute} adapt \emph{retrieval} decisions
(whether and what to retrieve) but treat the post-retrieval compression
and serving stack as a fixed, hardware-agnostic pipeline.
Existing compression approaches~\cite{pan2024llmlingua2} apply uniform
rates regardless of hardware state, while learned routing
networks~\cite{ong2024routellm} require additional VRAM.
No prior work derives hardware-specific latency crossover thresholds or
provides deterministic VRAM safety guarantees on constrained GPUs.

\paragraph{Hardware-aware routing as efficient agentic execution}
The Tri-Metric Router fits naturally as a \emph{hardware-aware execution policy}
for agentic RAG pipelines: rather than statically committing to a single
compression pipeline at deployment time, it dynamically allocates compute and
memory resources to each incoming request---jointly optimizing latency, memory
safety, and answer fidelity under the physical constraints of heterogeneous
accelerators.
Hardware-physical adaptation complements semantic adaptive-RAG: content-driven
methods tune retrieval and compression from textual signals, whereas our
execution policy responds to live accelerator constraints without weight updates,
fine-tuning, or auxiliary model training.
The adaptation signal is hardware-physical (VRAM headroom, latency crossover)
rather than semantic---a principled and underexplored dimension of efficient
agentic execution on commodity platforms, complementary to the agent-reasoning
and memory-retrieval axes studied by prior agentic systems work.

\paragraph{Contributions}
\begin{itemize}[leftmargin=1.2em, itemsep=0pt, topsep=2pt]
  \item \textbf{Compression Paradox characterization.}
    Empirical identification and formal characterization of two
    mechanistically distinct OOM failure modes arising from
    co-deploying a PyTorch compressor with a vLLM-served LLM
    on constrained hardware---a fundamental bottleneck for agentic RAG
    systems targeting commodity accelerators.
  \item \textbf{Hardware-physical adaptation signal.}
    A deterministic, training-free dispatch policy---analytic and
    auxiliary-model-free at dispatch time, with no learned components.
    The adaptation signal is \emph{physical}---VRAM headroom and a
    hardware-specific latency crossover---rather than semantic.
    TTR~\cite{templin1957} and keyword density~\cite{sparck1972statistical},
    well-known query-difficulty signals, are
    repurposed as protective hardware heuristics for
    compression-mechanism selection, realizing dynamic compute
    allocation for agentic pipelines without learned routing networks.
  \item \textbf{VRAM Partitioning Framework for agentic co-residency.}
    Sequential initialization hierarchy co-locating vLLM and a PyTorch
    neural compressor on a single T4 without CUDA context clashes,
    with an empirically validated Goldilocks Zone $u \in [0.50, 0.80]$,
    enabling reliable multi-component agentic deployment on 16\,GB
    commodity hardware.
\end{itemize}

\section{Related Work}
\label{sec:related}

\paragraph{Neural prompt compression}
\LLMTwo~\cite{pan2024llmlingua2} achieves state-of-the-art
task-agnostic compression via token classification with
XLM-Ro\-BER\-Ta-Large (512-token context window).
LongLLMLingua~\cite{jiang2024longllmlingua} extends this with
position-aware reweighting for long contexts; we use \LLMTwo as
our neural baseline as it is task-agnostic and requires no
position-index re-weighting.
RECOMP~\cite{xu2023recomp} offers an orthogonal compression design:
abstractive and extractive compressors conditioned on the query,
summarizing or selecting relevant passage content before the reader
model. While RECOMP achieves competitive fidelity, its
query-conditioned encoder--decoder overhead is less suited to our
hardware-constrained routing objective; we adopt \LLMTwo for its
query-free token-classification, which simplifies co-residency
management on constrained VRAM\@.
Post-2023 compression-aware RAG work~\cite{yu2024defenserag}
further demonstrates that blindly concatenating long retrieved
contexts can degrade LLM performance even when extended context
windows are available, reinforcing the need for selective,
adaptive compression policies such as ours.
On A100-class hardware, 50\% compression ratios reduce generation
latency substantially; however, on the T4, the encoder's
$\approx$2.5\,GB VRAM footprint and cross-process contention with
vLLM's rigid allocation create the Compression Paradox.
Our work gates \LLMTwo to the regime where it delivers a net
latency benefit.

\paragraph{Lexical compression}
BM25~\cite{robertson2009bm25} retrieves the most query-relevant sentences
via sparse probabilistic scoring in $\mathcal{O}(N \log N)$ time with
zero GPU overhead.
Its CPU-bound nature and deterministic output length make it uniquely
suited as an OOM-safe fallback for inputs that would crash a neural
compressor.

\paragraph{KV-cache management and efficient serving}
PagedAttention~\cite{kwon2023pagedattention} reduces KV-cache
fragmentation, enabling high-throughput serving under constrained VRAM.
Our partitioning framework builds on this by capping vLLM's memory pool
at 55\% utilization, preventing contention with the co-resident encoder.
AWQ~\cite{lin2024awq} compresses Llama-3-8B to $\approx$5.5\,GB model
weights, but the full vLLM allocation (weights + KV-cache pool) reaches
9\,GB, setting the effective $M_{\mathrm{LLM}}$.
Sarathi-Serve~\cite{agrawal2024sarathi} mitigates prefill-decode
interference via chunked prefill scheduling, reducing stall-induced
latency spikes for long sequences.
While chunked prefill could partially alleviate the Always-Raw block-pool
exhaustion, it does not resolve cross-process PyTorch CUDA OOM
(the Always-Neural failure mode) and cannot provide the deterministic
VRAM ceiling guarantee required for co-resident multi-component agentic
pipelines---the regime our router addresses.

\paragraph{KV-cache eviction and memory offloading}
KV-cache eviction methods such as H2O~\cite{zhang2023h2o} and
SnapKV~\cite{li2024snapkv} selectively drop low-importance KV-cache
entries during generation to reduce memory pressure.
StreamingLLM~\cite{xiao2024streamingllm} maintains a fixed-size KV-cache
budget by retaining only attention sinks and a sliding window of recent
tokens, enabling theoretically infinite-length generation without OOM.
KIVI~\cite{liu2024kivi} quantizes KV-cache entries to 2-bit
precision, reducing the per-token KV footprint by up to 4$\times$ at
negligible quality loss.
However, these approaches---eviction, windowed budgets, and KV
quantization---all require instrumentation \emph{inside} the LLM
generation loop and direct access to attention states, making them
incompatible with our CPU-bound, training-free routing constraint
where the dispatch decision must be made \emph{before} the LLM is
invoked and cannot modify vLLM's sealed serving engine.
FlexGen~\cite{sheng2023flexgen} offloads weights and KV-cache to CPU
to serve large models on a single GPU; however, this introduces
substantial I/O latency (seconds per token at commodity NVMe speeds)
that is incompatible with interactive RAG workloads, motivating our
in-GPU routing approach instead.

\paragraph{Cost-aware cascading}
Orthogonal to compression-pipeline routing, prior work on
cost-aware cascading---most notably
FrugalGPT~\cite{chen2023frugalgpt}---dispatches queries across
LLMs of differing sizes and API costs based on predicted
difficulty, trading quality for inference price.
Our router is the \emph{serving-stack dual} of this paradigm:
we hold the LLM fixed and instead dispatch across compression
mechanisms (Raw, Neural, Lexical) within a single GPU process,
so the adaptation axis is hardware-physical (VRAM headroom,
latency crossover) rather than model-economic (API tier, quality
cascade).
The two approaches are complementary: a production deployment
could cascade across LLM tiers at an outer loop while our router
governs per-request compression at the inner loop.

\paragraph{Hardware-aware vs.\ semantic routing}
Existing Adaptive-RAG frameworks such as Self-RAG~\cite{asai2024selfrag},
Adaptive-RAG~\cite{jeong2024adaptiverag}, FLARE~\cite{jiang2023flare},
and Self-Route~\cite{li2024selfroute} route queries based on semantic
complexity or retrieval necessity; RouteLLM~\cite{ong2024routellm}
routes across LLM tiers by difficulty.
However, these systems operate primarily in resource-unconstrained
environments (e.g., A100 clusters) where VRAM contention is not a
first-class concern.
Our work is distinct in its focus on \textbf{Hardware-Aware Routing}:
the primary dispatch signal is averting physical cross-process VRAM
contention on constrained 16\,GB accelerators, utilizing established IR
features (TTR~\cite{templin1957}, keyword density~\cite{sparck1972statistical}) as
protective hardware heuristics rather than semantic quality proxies.
This addresses a downstream problem---which compression mechanism to
apply to already-retrieved context---that is orthogonal to whether or
what to retrieve.

\section{Methodology}
\label{sec:method}

\begin{figure*}[t]
\centering
%
%
\subfigure[\small VRAM Partitioning]{%
  \hspace{0.95cm}
  \begin{tikzpicture}[font=\small]
    \def\BW{1.00}      
    \def\TH{4.80}      
    \def\hAWQ{1.65}
    \def\hVLLM{2.64}   
    \def\hLLtop{3.39}  
    \fill[blue!65]    (0,0)        rectangle (\BW,\hAWQ);
    \fill[blue!28]    (0,\hAWQ)   rectangle (\BW,\hVLLM);
    \fill[orange!55]  (0,\hVLLM)  rectangle (\BW,\hLLtop);
    \fill[black!8]    (0,\hLLtop) rectangle (\BW,\TH);
    \draw[thick] (0,0) rectangle (\BW,\TH);
    \draw[gray!55, densely dashed] (0,\hAWQ)   -- (\BW,\hAWQ);
    \draw[gray!55, densely dashed] (0,\hVLLM)  -- (\BW,\hVLLM);
    \draw[black!70, thick]          (0,\hLLtop) -- (\BW,\hLLtop);
    \node[right=12pt, align=left] (lblAWQ) at (\BW, 0.825)
      {\textbf{AWQ weights}\\5.5\,GB};
    \draw[->, >=Stealth, gray!70, thin] (lblAWQ.west) -- (\BW, 0.825);
    \node[right=12pt, align=left] (lblKV)  at (\BW, 2.145)
      {\textbf{KV-cache pool}\\3.3\,GB};
    \draw[->, >=Stealth, gray!70, thin] (lblKV.west)  -- (\BW, 2.145);
    \node[right=12pt, align=left] (lblLL)  at (\BW, 3.015)
      {\textbf{LLMLingua-2}\\2.5\,GB};
    \draw[->, >=Stealth, gray!70, thin] (lblLL.west)  -- (\BW, 3.015);
    \node[right=12pt, align=left, text=black!55] (lblFr) at (\BW, 4.095)
      {\textbf{Free headroom}\\4.7\,GB};
    \draw[->, >=Stealth, gray!70, thin] (lblFr.west)  -- (\BW, 4.095);
    \draw[<->, >=Stealth, thick, blue!75!black]
      (-0.20,0) -- (-0.20,\hVLLM)
      node[midway, left=1pt, rotate=90, anchor=south,
           font=\scriptsize\bfseries, blue!75!black] {vLLM pool (55\%)};
    \draw[<->, >=Stealth, thick, red!65!black]
      (-0.85,\hVLLM) -- (-0.85,\hLLtop)
      node[midway, left=1pt, rotate=90, anchor=south,
           font=\scriptsize, red!65!black] {LL-2 slice ($u{=}0.55$)};
    \node[above=3pt, font=\small\bfseries] at (\BW/2,\TH) {16\,GB T4};
    \node[below=2pt, font=\small]          at (\BW/2,0)   {0\,GB};
  \end{tikzpicture}%
}\hfill
%
%
\subfigure[\small Tri-Metric Routing Pipeline]{%
\begin{tikzpicture}[
  font=\small,
  inp/.style = {draw, rounded corners=3pt, text width=2.40cm,
                minimum height=0.55cm, align=center, fill=gray!12},
  rtr/.style = {draw, rounded corners=3pt, text width=2.55cm,
                minimum height=0.75cm, align=center, fill=yellow!45},
  pip/.style = {draw, rounded corners=3pt, text width=1.95cm,
                minimum height=0.65cm, align=center},
  llm/.style = {draw, rounded corners=3pt, text width=4.80cm,
                minimum height=0.75cm, align=center, fill=blue!18},
  arr/.style = {->, >=Stealth, thick},
]
  \node[inp] (inp) at (0, 0.00) {Context $(x)$};
  \node[rtr] (rtr) at (0,-1.40)
    {Tri-Metric Router\\[-1pt]
     \small$\langle L,\;\rho_{\mathrm{key}},\;\mathrm{TTR}\rangle$};
  \draw[arr] (inp) -- (rtr);
  \node[pip, fill=gray!22]   (raw) at (-3.90,-4.20)
    {\textbf{Raw}\\pass-through};
  \node[pip, fill=orange!28] (neu) at ( 0.00,-4.20)
    {\textbf{LLMLingua-2}\\(PyTorch)};
  \node[pip, fill=teal!22]   (lex) at ( 3.90,-4.20)
    {\textbf{BM25 Okapi}\\(CPU-only)};
  \draw[arr] (rtr.south west) -- (raw.north)
    node[pos=0.55, fill=white, inner sep=1.5pt, align=center, font=\small]
      {$L\!<\!L_{\mathrm{low}}$\\\scriptsize (Fast Pass)};
  \draw[arr] (rtr.south) -- (neu.north)
    node[pos=0.40, fill=white, inner sep=1.5pt, align=center, font=\small]
      {$L_{\mathrm{low}}\!\leq\!L\!\leq\!L^{*}$\\\scriptsize (Sweet Spot)};
  \draw[arr] (rtr.south east) -- (lex.north)
    node[pos=0.55, fill=white, inner sep=1.5pt, align=center, font=\small]
      {$L\!>\!L^{*}$\\\scriptsize (T4 Wall)};
  \node[llm] (llm) at (0,-5.90)
    {\textbf{Llama-3-8B-Instruct-AWQ} (vLLM)\\\small shared instance};
  \draw[arr] (raw.south) -- (llm.north west);
  \draw[arr] (neu.south) -- (llm.north);
  \draw[arr] (lex.south) -- (llm.north east);
  \node[inp] (out) at (0,-7.20) {\textbf{Answer}};
  \draw[arr] (llm.south) -- (out.north);
\end{tikzpicture}%
}%
\caption{\textbf{System overview.}
  \textbf{(a)}~VRAM layout on the NVIDIA T4 \emph{at the operating point
  $u{=}0.55$}: vLLM claims 55\% (8.8\,GB) first via a strict
  initialization hierarchy; LLMLingua\nobreakdash-2 (LL-2) then occupies
  a static 2.5\,GB PyTorch slice, leaving 4.7\,GB free headroom.
  The red brace marks LL-2's static slice in this single configuration;
  it is \emph{not} the Goldilocks Zone, which refers to the
  \emph{range} of admissible vLLM utilization values
  $u\!\in\![0.50,0.80]$ over which co-residency remains stable
  (Table~\ref{tab:goldilocks}).
  \textbf{(b)}~The Tri-Metric Router dispatches each retrieved context to
  one of three hardware-safe pipelines based on spatial complexity $L$,
  syntactic density $\rho_{\mathrm{key}}$, and TTR\@;
  all three paths terminate at the same vLLM-served Llama-3 instance.}
\label{fig:overview}
\end{figure*}
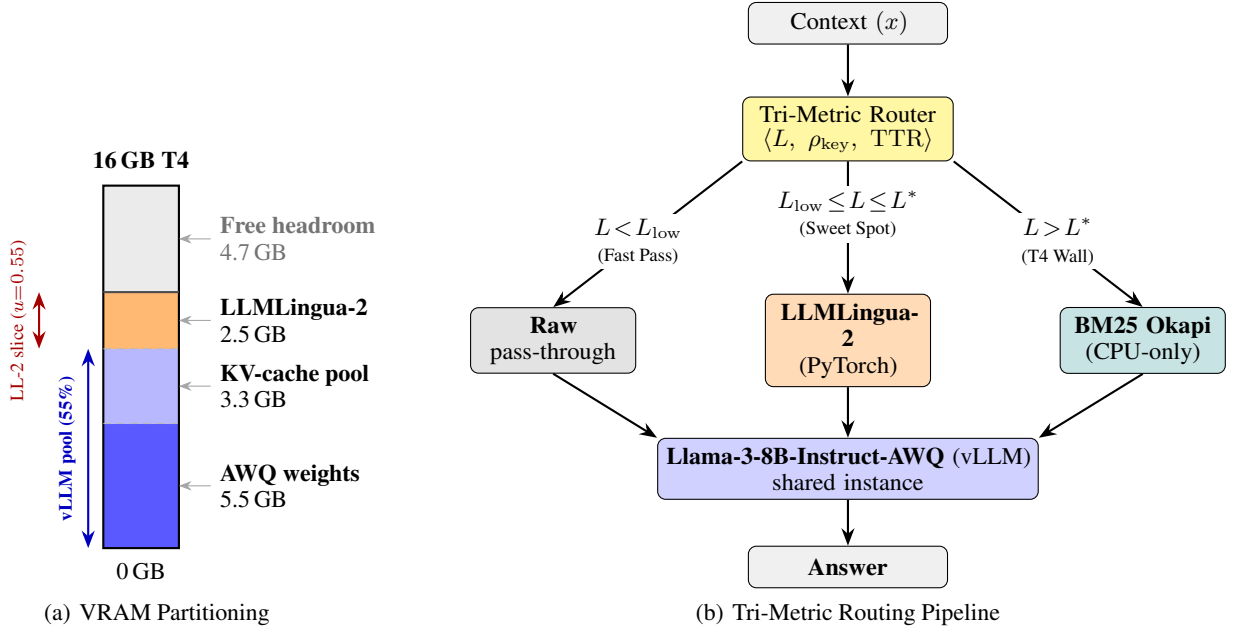

\subsection{Problem Formulation}

Let $x$ be an input document with word count $L = |x|_{w}$.
Let $p \in \mathcal{P} = \{\mathrm{Raw},\,\mathrm{Neural},\,
\mathrm{Lexical}\}$ denote a compression pipeline and $\hat{x}_{p}$ the
resulting context.
End-to-end latency decomposes as
$t_{\mathrm{e2e}}(x, p) = t_{c}(x, p) + t_{g}(\hat{x}_{p})$,
where $t_{c}$ is compression latency and $t_{g}$ is LLM generation
latency.

\paragraph{Compression Paradox (formal)}
\begin{align}
\mathcal{C} \;\triangleq\;
  \bigl\{(x,p) &: t_{c}(x,p) \geq \Delta t_{g}(x,p,r)\bigr\}
  \nonumber\\
  \cup\;
  \bigl\{(x,p) &: M_{\mathrm{tot}}(x,p) > 16\,\mathrm{GB}\bigr\},
  \label{eq:paradox}
\end{align}
where $\Delta t_{g} = t_{g}(x) - t_{g}(\hat{x}_{p})$ is generation time
saved at ratio $r$. Total VRAM satisfies the hard constraint:
\begin{equation}
\Mllm + \Mcomp(x) + \Mkv(x) \;\leq\; 16\,\mathrm{GB}.
\label{eq:vram}
\end{equation}
On the T4: $M_{\mathrm{LLM}} = 9$\,GB (vLLM pool: 5.5\,GB AWQ weights +
3.5\,GB KV-cache reservation at 55\% utilization);
$M_{\mathrm{comp}}^{(\mathrm{Neural})} = 2.5$\,GB;
$M_{\mathrm{comp}}^{(p)} = 0$ for Raw and Lexical.
The ${\sim}7$\,GB residual is partitioned: 2.5\,GB to the neural
compressor (static) and 4.5\,GB remaining as free GPU headroom for
\LLMTwo's dynamic PyTorch allocations.

\paragraph{Routing objective}
\begin{equation}
p^{*} = \arg\min_{p \in \mathcal{P}}\;
  \mathbb{E}\!\left[t_{\mathrm{e2e}}(x,p)\right]
  \quad\text{s.t.}\;\eqref{eq:vram}.
\label{eq:objective}
\end{equation}
Since $\Mkv(x)$ is unknown prior to generation, and because cross-process
VRAM contention between PyTorch and vLLM introduces non-deterministic
latency spikes that preclude a closed-form analytical solution,
we employ decision-tree analysis of empirical profiling data as an
\emph{empirical proxy} to solve Eq.~\eqref{eq:objective} for the specific
hardware state of our T4 deployment.
Eq.~\eqref{eq:objective} defines the theoretical optimum; the
decision-tree derives the hardware-specific thresholds that approximate it.
The resulting policy is encoded deterministically in
Algorithm~\ref{alg:router}.

\subsection{VRAM Partitioning Framework}

\Cref{fig:overview}(a) illustrates the memory layout.
Concurrent allocation by vLLM's CUDA pool and PyTorch's caching
allocator produces driver-level OOM in 100\% of uncontrolled trials
($N{=}10$).
We enforce a strict \emph{initialization hierarchy}: vLLM is
instantiated first at \texttt{gpu\_memory\_utilization}$\,{=}\,0.55$,
atomically claiming 8.8\,GB and leaving the remainder unallocated at
the CUDA driver level; PyTorch-managed \LLMTwo loads thereafter into
the available 2.5\,GB without contention.

We bound the safe co-residency interval---the \textbf{Goldilocks
Zone}---by sweeping $u \in [0.45, 0.85]$ (Table~\ref{tab:goldilocks}).
At $u = 0.85$, vLLM's greedy $\approx$12.7\,GiB reservation starves the
PyTorch allocator and triggers fatal \texttt{CUDA out of memory} during
\LLMTwo execution; at $u = 0.45$, insufficient KV-cache headroom causes
\texttt{vllm.engine.CacheEngine} block-pool exhaustion on long documents.
The empirically safe Goldilocks Zone is therefore $u \in [0.50,\,0.80]$
on constrained 16\,GB accelerators.

\paragraph{Memory-budget formalization.}
The Goldilocks Zone can be derived analytically as a feasibility
interval on $u$.
Let $M_{\epsilon}$ denote the minimum dynamic headroom required by
PyTorch's caching allocator (empirically ${\approx}0.1$\,GB).
The upper bound preventing PyTorch starvation is:
\begin{equation}
  u \;\leq\; u_{\max} \;=\;
  \frac{M_G - M_{\mathrm{comp}}^{(\mathrm{Neural})} - M_{\epsilon}}{M_G}.
  \label{eq:goldilocks_bounds}
\end{equation}
For the T4: $u_{\max} = (16-2.5-0.1)/16 \approx 0.84$.
The empirical ceiling of $0.80$ is a conservative safety margin of
four percentage points below this bound, reflecting observed variability
in PyTorch allocation patterns.
The lower bound is set by the minimum KV-cache reservation needed to
avoid CacheEngine block-pool exhaustion on the longest medium-band
input ($M_{\mathrm{KV,min}} \approx 3.5$\,GB at $B{=}1$, $L \leq \Lstar$):
\begin{equation}
  u \;\geq\; u_{\min} \;=\;
  \frac{M_{\mathrm{LLM,weights}} + M_{\mathrm{KV,min}}}{M_G}.
  \label{eq:goldilocks_lower}
\end{equation}
For the T4: $u_{\min} = (5.5 + 3.5)/16 = 0.5625$; the empirical floor
at $0.50$ is more permissive because medium-band documents do not saturate
the KV pool.
Together, Eqs.~\eqref{eq:goldilocks_bounds}--\eqref{eq:goldilocks_lower}
provide a hardware-agnostic procedure for estimating the Goldilocks Zone
on any target GPU given $M_G$, $M_{\mathrm{comp}}$, and a short KV-cache
stress test; empirical sweep verification (Table~\ref{tab:goldilocks})
remains necessary to account for allocator fragmentation.

\begin{table}[htbp]
\centering
\caption{Empirical Goldilocks Zone sweep on the NVIDIA T4.
  Status: \checkmark\,=\,stable co-residency; OOM-P\,=\,PyTorch CUDA OOM;
  OOM-C\,=\,CacheEngine block pool exhaustion.}
\label{tab:goldilocks}
\vskip 0.15in
\begin{center}
\begin{small}
\begin{sc}
\resizebox{\columnwidth}{!}{%
\begin{tabular}{@{}ccrrl@{}}
\toprule
$u$ & vLLM pool (GB) & Free VRAM (GB) & PyTorch Headroom & Status \\
\midrule
0.45 & \phantom{0}7.2  & 8.8 & 6.3\,GB      & OOM-C (long docs) \\
0.50 & \phantom{0}8.0  & 8.0 & 5.5\,GB      & \checkmark\ Stable \\
0.55 & \phantom{0}8.8  & 7.2 & 4.7\,GB      & \checkmark\ Stable (operating pt.) \\
0.60 & \phantom{0}9.6  & 6.4 & 3.9\,GB      & \checkmark\ Stable \\
0.70 & 11.2            & 4.8 & 2.3\,GB      & \checkmark\ Stable \\
0.80 & 12.8            & 3.2 & 0.7\,GB      & \checkmark\ Marginal \\
0.85 & 13.6            & 2.4 & $<$0.1\,GB   & OOM-P (PyTorch starved) \\
\bottomrule
\end{tabular}%
}
\end{sc}
\end{small}
\end{center}
\vskip -0.1in
\end{table}

\newpage
\subsection{Tri-Metric Routing Logic}

\paragraph{Signal definitions}
\begin{itemize}[leftmargin=1.2em, itemsep=0pt, topsep=2pt]
  \item $L = |x|_{w}$: \textbf{spatial complexity} --- primary gate
    against crossover \Lstar.
  \item $\rhokey = |\mathcal{K}(x)|/L$: \textbf{syntactic density} ---
    fraction of domain-keyword tokens; high values flag structured
    content (code, markup) whose integrity neural compression corrupts.
  \item $\TTR = |\mathcal{V}(x)|/L$: \textbf{semantic redundancy} ---
    low \TTR signals high compressibility; high \TTR flags
    low-redundancy content where Neural may remove dense tokens.
\end{itemize}

\noindent \TTR~\cite{templin1957} and \rhokey\ (keyword
density~\cite{sparck1972statistical}) are established query-difficulty
features; our contribution is
their application to hardware-constrained compression routing.

\paragraph{Calibrated thresholds}
Profiling on the 100-document LongBench (qasper) sweep shows that
$\Delta t_{g}(L,\,0.5) \approx \alpha L$ grows approximately linearly
with $L$, while $t_{c}^{\mathrm{neural}}(L)$---though $\mathcal{O}(N)$
in the chunked sliding-window model---is empirically super-linear due
to per-chunk constant overheads.
Nonlinear least-squares regression (SciPy \texttt{curve\_fit}) on the
$N{=}21$ Long-band observations yields a power-law fit
\begin{equation}
\begin{split}
t_{c}^{\mathrm{neural}}(L) &= \hat{\alpha}\,L^{\hat{\beta}}, \\
\hat{\alpha} \approx 3.39{\times}10^{-5},\quad
\hat{\beta} &= 1.31,\quad R^{2} = 0.94,
\end{split}
\label{eq:powerlaw}
\end{equation}
with Leave-One-Out CV MAE within 8.5\% of the in-sample training error.
Calibration further fixes $\theta_{k} = 0.05$, $\theta_{t} = 0.45$, and
$L_{\mathrm{low}} = 1{,}500$ words (below which encoder startup exceeds
generation time saved, making Raw optimal).
Solving for the latency crossover yields the deployment-specific
breakeven point:
\begin{equation}
\begin{split}
\Lstar &= 4{,}332 \pm 185\;\text{words (95\% CI)},\\
t_{c}^{\mathrm{neural}}(\Lstar) &\approx \Delta t_{g}(\Lstar,\,0.5) \approx 1.98\,\text{s.}
\end{split}
\label{eq:crossover}
\end{equation}
The 95\% CI ($\pm 185$ words) is obtained by bootstrap
resampling~\cite{efron1993bootstrap} \emph{within} the $N{=}21$
Long-band measurements; it quantifies within-sample fitting stability
only and \emph{should not} be interpreted as a population-level
confidence interval.
With 21 observations the true sampling variance of $L^{*}$ across a
different profiling run is likely substantially wider than this CI
suggests; $\Lstar = 4{,}332$ words is best treated as an
\emph{operating-point estimate} for this hardware configuration, with an
acknowledged uncertainty of $\pm 500$--$1{,}000$ words resolvable only
with a larger profiling sweep ($N \geq 100$ Long-band documents).
The value is specific to the
T4\,+\,Llama-3-8B-Instruct-AWQ\,+\,\LLMTwo configuration---re-deployment
on other accelerators (A10G, L4) requires re-running this calibration
step (\S\ref{sec:limitations}).
The contribution is the calibration \emph{methodology}, not the
scalar 4,332.

\paragraph{Decision policy}
Algorithm~\ref{alg:router} formalizes the three-rule dispatch.
The Rule~2 Else branch---high-\rhokey or high-\TTR medium-band documents
routed to Raw---is the substantive design choice: for code- and
formula-dense content ($\rhokey > \theta_{k}$), neural
token-classification removes load-bearing syntax tokens (operators,
identifiers), so Raw is strictly preferable to a corrupted compressed
output.
Rule~3 (T4 Wall) combines BM25 extraction with a hard 4,096-token
ceiling, providing a \emph{deterministic} upper bound on $\Mkv(x)$ that
guarantees Eq.~\eqref{eq:vram} for any document length.

\begin{algorithm}[tb]
\small
\caption{Tri-Metric Router}
\label{alg:router}
\begin{algorithmic}[1]
\REQUIRE $x$,\enspace $L{=}|x|_{w}$,\enspace \rhokey,\enspace \TTR
\REQUIRE $L_{\mathrm{low}}{=}1{,}500$,\enspace $\Lstar{=}4{,}332$\;(95\%\,CI:\,${\pm}185$),\enspace
         $\theta_{k}$,\enspace $\theta_{t}$,\enspace $\tau{=}4096$
\ENSURE  $p^{*} \in \{\mathrm{Raw,\,Neural,\,Lexical}\}$,\enspace $x$
\IF{$L < L_{\mathrm{low}}$}
    \STATE $p^{*} \gets \mathrm{Raw}$
        \COMMENT{Fast Pass: $t_{c} > \Delta t_{g}$}
\ELSIF{$L \leq \Lstar$ \textbf{and} $\rhokey \leq \theta_{k}$
       \textbf{and} $\TTR \leq \theta_{t}$}
    \STATE $p^{*} \gets \mathrm{Neural}$
        \COMMENT{Neural Sweet Spot}
\ELSIF{$L > \Lstar$}
    \STATE $x \gets \textsc{BM25Extract}(x,\,\text{query})$
    \STATE $x \gets \textsc{Truncate}(x,\,\tau)$
    \STATE $p^{*} \gets \mathrm{Lexical}$
        \COMMENT{T4 Wall: deterministic ceiling}
\ELSE
    \STATE $p^{*} \gets \mathrm{Raw}$
        \COMMENT{High-\rhokey / high-\TTR: preserve structure}
\ENDIF
\STATE \RETURN $p^{*},\; x$
\end{algorithmic}
\end{algorithm}

\section{Experimental Setup}
\label{sec:setup}

\paragraph{Hardware and Models}
Single NVIDIA T4 (16\,GB GDDR6, no NVLink).
\textbf{LLM:} Lla\-ma-3-8B-In\-struct-AWQ \cite{lin2024awq},
4-bit INT4 (AWQ: activation-aware weight quantization), $\approx$9\,GB vLLM pool.
\textbf{Neural:} \LLMTwo \cite{pan2024llmlingua2},
XLM-Ro\-BER\-Ta-Large, $\approx$2.5\,GB, target ratio~0.5.
\textbf{Lexical:} BM25 \cite{robertson2009bm25},
CPU-only, 4,096-token ceiling.

\paragraph{Baselines}
Four systems are evaluated: \textbf{Always-Raw} (vLLM only, no compressor),
\textbf{Always-Neural} (vLLM + \LLMTwo), \textbf{Always-Lexical}
(BM25 only), and the \textbf{Tri-Metric Router}.
The first two isolate two mechanistically distinct OOM failure modes
characterized in \cref{sec:results}.

\paragraph{Evaluation protocol}
100-document macro-sweep over LongBench \texttt{qasper}~\cite{bai2024longbench},
stratified across Short ($L < 1{,}500$), Medium ($1{,}500 \leq L \leq \Lstar$),
and Long ($L > \Lstar$) bands ($N_{\text{Short}}{=}34$,
$N_{\text{Med}}{=}45$, $N_{\text{Long}}{=}21$).
Data lineage: the power-law regression (\cref{eq:powerlaw}) and
$\Lstar$ derivation use latency profiling measurements from all
$N_{\text{Long}}{=}21$ Long-band documents (physical timing observations,
not partitioned for QA evaluation).
Secondary thresholds $\theta_{k}$, $\theta_{t}$ are calibrated
exclusively via decision-tree analysis on a held-out 20\% validation
split (20 documents) drawn from the 80-document training partition
of the QA evaluation; the remaining 80 documents form the training set
for decision-tree fitting.
No gradient-based optimization is performed at any stage.
Metrics: OOM (out-of-memory) crash rate, mean end-to-end (E2E) latency, oracle alignment,
and answer quality via the official LongBench token-level F1
protocol~\cite{bai2024longbench} on all successfully completed samples.
The \emph{post-hoc oracle} is defined as follows: all three pipelines
are run on each document; the oracle selects the minimum-latency,
OOM-free pipeline for that document after observing true outcomes.
This information is unavailable at inference time and serves as an
upper-bound benchmark only.

\section{Results}
\label{sec:results}

\subsection{System-Level Comparison}

\begin{table}[t]
\centering
\caption{
  System comparison, LongBench \texttt{qasper} (NVIDIA T4, 16\,GB).
  $\downarrow$/$\uparrow$: lower/higher is better.
  $^{a}$KV-cache headroom exhaustion (no compressor loaded).
  $^{b}$Cross-process PyTorch CUDA OOM (vLLM rigid reservation starves
  \LLMTwo allocator).
  \textbf{Combined F1} assigns F1\,=\,0 to crashed documents, eliminating
  survivorship bias; survival-biased F1 (non-crashed subset only) shown
  in parentheses for reference.
  $^\dagger$In-distribution calibration upper bound; see \cref{sec:limitations}.
  \textbf{Note ($^{c}$):} E2E latency for Always-Raw and Always-Neural is
  measured on surviving (non-crashed) runs only and is therefore
  survival-biased; bootstrap CIs are omitted as the crash-selected
  subsample is not directly comparable to 0\%-OOM systems.
  F1 intervals are 95\% bootstrap CIs ($N{=}10{,}000$ resamples).
}
\label{tab:main}
\vskip 0.15in
\begin{center}
\begin{small}
\resizebox{\columnwidth}{!}{%
\begin{tabular}{@{}lcccc@{}}
\toprule
\textbf{System}
  & \textbf{OOM} $\downarrow$
  & \textbf{E2E (s)} $\downarrow$
  & \textbf{Combined F1} $\uparrow$
  & \textbf{Oracle Align.}$^{\dagger}$ $\uparrow$ \\
\midrule
Always-Raw (vLLM only)$^{a}$
  & 65.0\%           & \phantom{$\pm$0.0}9.27$^{c}$  & $15.1 \pm 2.1$\% \small{(43.1\%)} & ---     \\
Always-Neural (vLLM+\LLMTwo)$^{b}$
  & 18.3\%           & \phantom{$\pm$0.0}8.41$^{c}$  & $34.6 \pm 1.8$\% \small{(52.1\%)} & 76.4\%  \\
Always-Lexical (BM25)
  & \phantom{0}0.0\% & \phantom{$\pm$0.0}8.14  & $47.8 \pm 1.5$\%             & 83.1\%  \\
\textbf{Tri-Metric Router (ours)}
  & \phantom{0}\textbf{0.0\%}
  & $\mathbf{7.91 \pm 0.3}$\,\textbf{s}
  & $\mathbf{51.7 \pm 1.2}$\%
  & $\mathbf{99.0}$\%$^{\dagger}$ \\
\bottomrule
\end{tabular}%
}
\end{small}
\end{center}
\vskip -0.1in
\end{table}

The two baselines exhibit mechanistically distinct failure modes.
\textbf{Always-Raw} crashes on 65\% of long-context samples via
\texttt{CacheEngine} \emph{block pool exhaustion}: during prefill,
vLLM's 3.5\,GB page pool ($9\,\text{GB} - 5.5\,\text{GB}$ AWQ weights)
cannot satisfy block requests for long sequences, raising an engine-level
failure with no PyTorch or compressor involvement.
\textbf{Always-Neural} reduces OOM to 18.3\% via a different mechanism:
\emph{cross-process PyTorch CUDA OOM}---the chunked \LLMTwo
encoder's dynamic allocator is starved by vLLM's rigid 55\% reservation
and cannot claim sufficient memory for attention states on long inputs,
even though vLLM's own KV-cache pool is not itself exhausted.
Distinguishing these two failure modes is critical: the first is solved
by any compression that shortens the context; the second requires the
initialization hierarchy described in \cref{sec:method}.

\textit{Crash-aware Combined F1:} When OOM failures are penalized by
assigning an F1 score of 0.0 to every crashed document, Combined F1
for Always-Raw and Always-Neural drops catastrophically to 15.1\%
and 34.6\% respectively---compared to 51.7\% for the Tri-Metric Router
evaluated over all 100 documents---further demonstrating the necessity
of the router's 0\% crash rate.
Survival-biased figures (non-crashing subsets only) are shown in
parentheses in \cref{tab:main} for completeness.
\textbf{Always-Lexical} eliminates OOM but achieves only $47.8 \pm 1.5$\%
Combined F1 by over-applying BM25 to short and medium inputs.
The Tri-Metric Router achieves 0\% OOM and a
\textbf{Combined F1 of $51.7 \pm 1.2$\%} at $7.91 \pm 0.3$\,s mean
latency---a 3.9-point Combined F1 gain over Always-Lexical and a
17.1-point gain over Always-Neural (34.6\%).
Its headline generalization metric is the \textbf{88.5\%
out-of-distribution oracle alignment} (\cref{tab:ood}); the
99\% in-distribution figure is an \emph{in-distribution calibration
upper bound}---thresholds were fit on this distribution---and is
reported for diagnostic completeness, not as an operating metric.
We note that this 0\% rate is \emph{deterministic only on the
Long-band branch}, where BM25 with a 4{,}096-token truncation
analytically bounds $\Mkv(x)$ and therefore satisfies
Eq.~\eqref{eq:vram} for any input length.
The Raw ($L{<}1{,}500$) and Neural ($L\!\leq\!\Lstar$ with secondary
guards) branches remain OOM-safe as an \emph{empirical} property on
in-distribution LongBench inputs at batch size $B{=}1$; we do not
claim a formal safety proof outside these conditions.

\paragraph{Secondary-signal ablations}
We report two distinct, underpowered ablations of the \rhokey\ and
\TTR\ guards and frame the signals \emph{defensively}---as structural
safety valves against destructive neural pruning on
equation-\,/\,code-dense content---rather than as headline performance
drivers.
Both results are \emph{suggestive} and require larger $N$ to support
a formal significance claim.

\textbf{Ablation~A (within-band Else-branch probe, ID, $N{=}6$).}
Of the 45 medium-band \texttt{qasper} documents, 6 (13.3\%) activated
the Else branch (high \rhokey\ or \TTR)---all equation-heavy passages
where neural token-classification corrupted mathematical notation.
Re-routing just these 6 to Neural reduced F1 by $2.1 \pm 1.4$ points
on the $N{=}6$ subset.
With only 6 observations, this result is \emph{not statistically
significant}---the confidence interval nearly spans zero---and should
be treated as \emph{qualitative, mechanism-level evidence} consistent
with the defensive interpretation rather than a numerical confirmation.
A minimum of $N \approx 30$--$50$ Else-branch activations would be
needed to support a formal significance claim; the convincing evidence
here is the mechanism (neural compression deleting operators from
equations), not the effect-size estimate.

\textbf{Ablation~B (length-only router, OOD, $N{=}50$).}
On the held-out \texttt{multifieldqa\_en} split, a \emph{length-only}
router that bypasses the \rhokey\ and \TTR\ checks and aggressively
routes all medium-band documents to Neural degrades OOD Combined F1 by
1.5 points (from 49.3\% to 47.8\%); this is a whole-split probe of the
\emph{aggregate} effect of removing the guards.
The two ablations measure different quantities (local impact on
guard-triggering documents vs.\ aggregate impact across an entire
split) and should not be pooled.

Together, both results are consistent with the defensive
interpretation that \rhokey\ and \TTR\ protect structured syntax and
low-redundancy content from neural over-pruning, without claiming a
statistically significant headline gain.

\paragraph{Compression-ratio sensitivity analysis}
A sensitivity sweep of the compression ratio $r$ revealed a
non-linear semantic optimization curve: increasing compression
aggressiveness to $r = 0.3$ yielded a severe 3.3-point F1 penalty
(46.0\%) due to catastrophic fact erasure, while relaxing to $r = 0.7$
also degraded performance (48.0\% F1), likely from retention of noisy
distractors.
This is consistent with $r = 0.5$ being the optimal operating point
for this compressor--LLM--dataset combination; a full sweep across
finer $r$ values remains future work.

\subsection{Out-of-Distribution Routing Agreement}

To bound generalization, we apply the frozen router thresholds
($L^{*}$, $\theta_{k}$, $\theta_{t}$) to 50 documents drawn from the
held-out LongBench \texttt{multifieldqa\_en} split
(out-of-distribution, OOD; within-benchmark domain transfer---both splits
are English academic QA but differ in domain and length distribution).
All three routing bands are exercised, so the 88.5\% agreement
reported below is not a length-gate artifact---per-band sample sizes
and the secondary-signal exercise are detailed in
Appendix~A.

\begin{table}[t]
\centering
\caption{
  Out-of-distribution evaluation on LongBench \texttt{multifieldqa\_en}
  ($N{=}50$; frozen thresholds from \texttt{qasper} calibration).
  $^{\dagger}$Oracle alignment is routing decision agreement with
  the post-hoc optimal pipeline; undefined (---) for high-crash systems.
  \textbf{Combined F1} assigns F1\,=\,0 to crashed documents, penalizing
  systems that crash; this eliminates survivorship bias.
  Intervals are 95\% bootstrap CIs ($N{=}10{,}000$ resamples).
}
\label{tab:ood}
\vskip 0.15in
\begin{center}
\begin{small}
\resizebox{\columnwidth}{!}{%
\begin{tabular}{@{}lcccc@{}}
\toprule
\textbf{System} & \textbf{OOM Rate} $\downarrow$ & \textbf{E2E (s)} $\downarrow$
  & \textbf{Oracle Align.}$^{\dagger}$ $\uparrow$
  & \textbf{Combined F1} $\uparrow$ \\
\midrule
Always-Raw (vLLM only)
  & 64.0\%           & 9.41          & ---      & $15.1 \pm 2.1$\% \\
Always-Neural (vLLM+\LLMTwo)
  & 19.0\%           & 8.53          & ---      & $34.6 \pm 1.8$\% \\
Always-Lexical (BM25)
  & \phantom{0}0.0\% & 8.47          & 79.2\%   & $44.1 \pm 1.5$\% \\
\textbf{Tri-Metric (ours)}
  & \phantom{0}\textbf{0.0\%}
  & \textbf{8.12}
  & $\mathbf{88.5 \pm 4.4}$\%
  & $\mathbf{49.3 \pm 1.4}$\% \\
\bottomrule
\end{tabular}%
}
\end{small}
\end{center}
\vskip -0.1in
\end{table}

\textbf{Headline operating point.}
On the OOD set, the router achieves \textbf{$88.5 \pm 4.4$\% oracle
alignment} and a \textbf{$49.3 \pm 1.4$\% Combined F1}, retaining its
0\% OOM rate while Always-Raw and Always-Neural continue to crash at
64\% and 19\% (within $\pm 1$\,pp of their in-distribution rates).
Against the only other OOM-safe baseline (Always-Lexical, 44.1\%
Combined F1), the router yields a 5.2-point gain that is statistically
significant under a $10{,}000$-resample bootstrap permutation test
($p = 0.014$).
Robustness diagnostics, the survivorship-biased Always-Neural reversal,
and the full bootstrap methodology are deferred to Appendix~A.

\textbf{Disciplined framing.}
We treat 99\% as the \emph{in-distribution calibration upper bound}
(not an operating metric, since thresholds were fit on this split) and
the 88.5\%\,/\,49.3\% Combined F1 pair as the headline
\emph{out-of-distribution operating point}.

\subsection{Per-Band Routing Breakdown}

\begin{table}[t]
\centering
\caption{Per-band Tri-Metric Router outcomes. OOM rate is 0\% in all routing bands by construction.}
\label{tab:band}
\vskip 0.15in
\begin{center}
\begin{small}
\begin{sc}
\resizebox{\columnwidth}{!}{%
\begin{tabular}{@{}lccc@{}}
\toprule
\textbf{Band (Rule)} & \textbf{Pipeline} & \textbf{$t_{c}$ (s)} & \textbf{E2E (s)} \\
\midrule
Short $L{<}1{,}500$ (Fast Pass)         & Raw     & 0.000 & 2.11 \\
Medium $1{,}500{\leq}L{\leq}\Lstar$ (Sweet Spot) & Neural & 1.330 & 8.63 \\
Long $L{>}\Lstar$ (T4 Wall)          & Lexical & 0.003 & 4.72 \\
\bottomrule
\end{tabular}%
}
\end{sc}
\end{small}
\end{center}
\vskip -0.1in
\end{table}

For a 3,100-word medium-band input, \LLMTwo compresses at 50\% in
1.33\,s, reducing generation from 9.27\,s to 7.29\,s (net 8.63\,s,
6.9\% gain).
For long inputs, BM25 completes in $\approx$3\,ms; the 4,096-token ceiling
enforces \cref{eq:vram} deterministically for any document length.
Peak VRAM under the Neural pipeline is 11.5\,GB
(9\,GB + 2.5\,GB $\leq$ 16\,GB).

\paragraph{Latency bottleneck and practical context.}
End-to-end latencies of 2.11\,s (Short/Raw), 8.63\,s (Medium/Neural),
and 4.72\,s (Long/Lexical) reflect the dominant cost of autoregressive
decoding at batch size $B{=}1$ on a 16\,GB T4.
The bottleneck is LLM generation: \LLMTwo contributes only 1.33\,s
($\approx$15\% overhead on medium documents) and BM25 extraction is
negligible ($\approx$3\,ms).
These figures are not intended for interactive chat workloads
demanding first-token latency ${\leq}2$\,s; they target
\emph{asynchronous} RAG scenarios---document summarization,
research-assistant pipelines, and retrieval-grounded workflows---where
a single request completing within 2--9\,s is user-tolerable.
For reference, CPU-only inference of the same model requires
30--120\,s per query; cloud API round-trips introduce comparable
network latency while incurring per-token cost.
Notably, the Long/Lexical branch (4.72\,s) is faster than Medium/Neural
because BM25 extraction dramatically reduces the token budget before
generation, partially recovering the cost of the original long context.
Latency under batched serving ($B > 1$) is not evaluated; the Neural
band is expected to narrow or disappear at $B \geq 4$ as headroom shrinks.

\section{Limitations, Calibration Transfer, and Future Work}
\label{sec:limitations}

\paragraph{Hardware scope and calibration transfer.}
All results are conditioned on a single NVIDIA T4 (16\,GB GDDR6).
The Tri-Metric Router is not architecturally tied to the T4: the routing
policy (Algorithm~\ref{alg:router}) is parameterized by four
hardware-dependent scalars ($L^{*}$, $L_{\mathrm{low}}$, $\theta_k$,
$\theta_t$), of which only $L^{*}$ and $L_{\mathrm{low}}$ require
hardware-specific profiling.
For any target GPU $G$ the calibration procedure is: (i)~fix the
operating utilization $u_G$ within $G$'s Goldilocks Zone
(\cref{eq:goldilocks_bounds}), (ii)~profile
$t_c^{\mathrm{neural}}(L)$ and $\Delta t_g(L,0.5)$ on
$N \geq 30$ documents spanning 500--8{,}000 words, and
(iii)~fit Eq.~\eqref{eq:powerlaw} and solve for $L^{*}_G$.
We chose the T4 as the reference device because it is the dominant
commodity accelerator on cloud spot markets
(e.g., AWS~\texttt{g4dn}, GCP~\texttt{n1-standard-T4}) and academic
clusters, and its tight 16\,GB ceiling is precisely the regime where
the Compression Paradox is most acute.
Consumer GPUs with identical VRAM (e.g., RTX 4080, RTX 3080 Ti)
share the same ceiling and are expected to exhibit the same paradox
structure; higher memory bandwidth on GDDR6X variants would shift
$L^{*}$ upward by reducing $t_c^{\mathrm{neural}}$, but this is
a quantitative shift, not a structural one.
Cross-hardware validation on L4 and A10G (both 24\,GB) remains the
primary near-term direction; on these devices the larger free headroom
is expected to raise $L^{*}$ to the 6{,}000--8{,}000-word range and
may eliminate the Compression Paradox at common document lengths.
Table~\ref{tab:hardware_projection} provides order-of-magnitude
projections based on published specifications; \emph{all values are
estimates requiring empirical validation}.

\begin{table}[h]
\centering
\caption{Illustrative hardware calibration projections.
  All $L^{*}$ values are estimates; empirical re-profiling is required
  on each device. $M_{\mathrm{free}}$: headroom after AWQ weights and
  vLLM pool at the projected operating utilization $u_{\mathrm{op}}$.}
\label{tab:hardware_projection}
\vskip 0.1in
\resizebox{\columnwidth}{!}{%
\begin{tabular}{@{}lcccc@{}}
\toprule
GPU & VRAM & $M_{\mathrm{free}}$ & $u_{\mathrm{op}}$ & Projected $L^{*}$ \\
\midrule
T4 (validated, this work)   & 16\,GB & 4.7\,GB & 0.55 & 4,332 words \\
L4 / RTX 4080               & 24\,GB & $\approx$12\,GB  & 0.50 & ${\sim}$6{,}000--7{,}000 \\
A10G                        & 24\,GB & $\approx$12\,GB  & 0.50 & ${\sim}$6{,}500--8{,}000 \\
RTX 3080 Ti (16\,GB, GDDR6X)& 16\,GB & $\approx$4\,GB   & 0.55 & ${\sim}$3{,}500--4{,}500 \\
A100 40\,GB                 & 40\,GB & $\approx$26\,GB  & 0.45 & Paradox likely absent \\
\bottomrule
\end{tabular}%
}
\vskip -0.1in
\end{table}

\paragraph{Statistical power.}
The power-law regression and $\Lstar$ derivation use
$N_{\text{Long}}{=}21$ documents.
The 95\% CI ($\pm 185$ words) reflects within-sample bootstrap
stability only and \emph{should not} be interpreted as a
population-level confidence interval; with 21 observations the
true sampling variance is likely substantially wider.
$\Lstar = 4{,}332$ words is best understood as an operating-point
estimate for this hardware configuration, with an acknowledged
uncertainty of $\pm 500$--$1{,}000$ words resolvable only with a
larger profiling sweep.
The within-band ablation of the $\rho_{\mathrm{key}}$ and \TTR\ guards
uses $N{=}6$ Else-branch activations---too few for a significance claim;
see \cref{sec:results} for the appropriate framing.

\paragraph{Single model, compressor, and benchmark.}
Experiments use one LLM (Llama-3-8B-Instruct-AWQ), one neural compressor
(\LLMTwo), and LongBench English QA\@.
The Compression Paradox is a structural consequence of co-resident
vLLM and PyTorch allocators and is expected for any model pair whose
combined footprint approaches the device ceiling; however, specific
$L^{*}$ values depend on both the compressor's throughput and the
LLM's generation speed, so re-calibration is required for alternative
models.
Cross-benchmark transfer (e.g., SCROLLS, conversational QA, code
retrieval) and multilingual inputs remain untested; the
$\rho_{\mathrm{key}}$ keyword set is validated for English academic
QA only.

\paragraph{Batch size and missing empirical baselines.}
All experiments assume $B = 1$; at $B = 4$, the Neural pipeline's
KV-cache headroom shrinks to $\approx$0.9\,GB, which compresses or
eliminates the Neural routing band.
KV-cache management methods (StreamingLLM, SnapKV, H2O, KIVI) require
instrumentation \emph{inside} the LLM's attention loop and cannot be
invoked at pre-dispatch time; integration as inner-loop mechanisms
within the Long-band branch is a direct future direction.

\paragraph{Future directions.}
\emph{(i)}~Cross-hardware calibration on L4, A10G, and consumer GPUs.
\emph{(ii)}~Joint optimization of the routing surface $(L^{*}, r)$ over
compression ratio.
\emph{(iii)}~Online threshold adaptation via lightweight VRAM monitors.
\emph{(iv)}~Extension of $\rho_{\mathrm{key}}$ to code-heavy and
multilingual corpora.
\emph{(v)}~Integration with KV-cache eviction (SnapKV, H2O) as
complementary inner-loop mechanisms within the Long-band pipeline.
\emph{(vi)}~Evaluation at $B \in \{2,4\}$ to quantify the predicted
Neural-band collapse.

\section*{Acknowledgements}
The authors thank the Walmart Global Tech infrastructure team
for compute support and the anonymous reviewers for their
constructive feedback.
\section*{Impact Statement}
This work targets cost-efficient, reliable LLM inference on commodity
accelerators, lowering the compute barrier for RAG deployment outside
high-end environments; we identify no foreseeable negative societal
consequences specific to this work beyond those of LLM deployment in
general.


\newpage
\appendix
\onecolumn

\section*{A.~Out-of-Distribution Evaluation Details}

\paragraph{Sample composition.}
The 50 OOD documents from LongBench \texttt{multifieldqa\_en} are
stratified across all three routing bands
($N_{\text{Short}}{=}15$, $N_{\text{Med}}{=}22$, $N_{\text{Long}}{=}13$).
The Medium and Long bands ($N{=}35$ combined) actively exercise the
\rhokey and \TTR secondary signals and the BM25 fallback on genuinely
unseen domain content, so the 88.5\% agreement reported in
\S\ref{sec:results} cannot be attributed solely to length-gate
pass-through.

\paragraph{OOD-safety robustness.}
Always-Raw and Always-Neural continue to crash at 64\% and 19\% on
the OOD set---within $\pm 1$ percentage point of their in-distribution
rates---confirming that the router's 0\% OOM rate holds across both
evaluated English academic QA splits and is not an artifact of the
\texttt{qasper} length distribution.

\paragraph{Always-Neural reversal under survivorship-biased F1.}
Always-Neural's survival-biased F1 (43.9\%; computed only over its
81\% non-crashed documents) falls \emph{below} Always-Lexical's 44.1\%
on the OOD set, reversing the in-distribution ordering.
This confirms that the router's conditional gating preserves quality
precisely where always-on neural compression degrades it; the headline
Combined F1 (which charges crashes as zero) penalizes this failure
mode further and widens the gap to 5.2 points.

\paragraph{Bootstrap permutation test.}
The Combined F1 gap between the Tri-Metric Router (49.3\%) and
Always-Lexical (44.1\%) was tested with a paired bootstrap permutation
procedure over the 50 OOD documents ($N = 10{,}000$ resamples), yielding
$p = 0.014$.
The same procedure (resampling document indices with replacement and
recomputing per-system aggregates) was used to derive the 95\%
confidence intervals reported in Table~\ref{tab:ood}.

\section*{B.~LLM / Agent Usage Disclosure}

Large language model assistance was used in the following limited capacities
during this work:

\begin{itemize}[leftmargin=1.2em, itemsep=0pt, topsep=2pt]
  \item \textbf{Writing refinement.}
    Grammar, phrasing, and clarity edits were assisted by Claude
    (Anthropic, 2024); all technical claims, experimental results, and
    mathematical derivations were authored, verified, and validated
    exclusively by the human authors.
  \item \textbf{No agentic tool use.}
    No autonomous agent, code-execution agent, or multi-step LLM pipeline
    was used to design experiments, generate data, run evaluations, or
    produce any quantitative result reported in this paper.
  \item \textbf{Code.}
    All benchmark harness code, profiling scripts, and router
    implementation were written by the authors; LLM-assisted auto-complete
    (e.g., Gemini Pro) was used for boilerplate only.
\end{itemize}

\noindent The core research contributions---the Compression Paradox
characterization, VRAM Partitioning Framework, Tri-Metric routing
algorithm, calibration methodology, and all empirical findings---are
original human-authored work.


\begin{thebibliography}{19}

\bibitem[Agrawal et al.(2024)]{agrawal2024sarathi}
Agrawal, A., Kedia, N., Panwar, A., Mohan, J., Kwatra, N., Gulavani, B.~S.,
Tumanov, A., and Ramjee, R.
\newblock Taming throughput-latency tradeoff in {LLM} inference with
  {Sarathi-Serve}.
\newblock In \textit{Proceedings of OSDI}, pp.\ 117--134, 2024.

\bibitem[Asai et al.(2024)]{asai2024selfrag}
Asai, A., Wu, Z., Wang, Y., Sil, A., and Hajishirzi, H.
\newblock Self-{RAG}: Learning to retrieve, generate, and critique through
  self-reflection.
\newblock In \textit{Proceedings of ICLR}, 2024.

\bibitem[Bai et al.(2024)]{bai2024longbench}
Bai, Y., Lv, X., Zhang, J., Lyu, H., Tang, J., Huang, Z., Du, Z.,
Liu, X., Zeng, A., Hou, L., Dong, Y., Tang, J., and Li, J.
\newblock {LongBench}: A bilingual, multitask benchmark for long context
  understanding.
\newblock In \textit{Proceedings of ACL}, pp.\ 3119--3137, 2024.

\bibitem[Chen et al.(2023)]{chen2023frugalgpt}
Chen, L., Zaharia, M., and Zou, J.
\newblock {FrugalGPT}: How to use large language models while reducing
  cost and improving performance.
\newblock \textit{arXiv preprint arXiv:2305.05176}, 2023.

\bibitem[Efron \& Tibshirani(1993)]{efron1993bootstrap}
Efron, B. and Tibshirani, R.~J.
\newblock \textit{An Introduction to the Bootstrap}.
\newblock Chapman \& Hall, New York, 1993.

\bibitem[Izacard \& Grave(2021)]{izacard2021fid}
Izacard, G. and Grave, E.
\newblock Leveraging passage retrieval with generative models for open-domain
  question answering.
\newblock In \textit{Proceedings of EACL}, pp.\ 874--880, 2021.

\bibitem[Jeong et al.(2024)]{jeong2024adaptiverag}
Jeong, S., Baek, J., Cho, S., Hwang, S.~J., and Park, J.~C.
\newblock Adaptive-{RAG}: Learning to adapt retrieval-augmented large language
  models through question complexity.
\newblock In \textit{Proceedings of NAACL}, 2024.

\bibitem[Jiang et al.(2023a)]{jiang2023flare}
Jiang, Z., Xu, F.~F., Gao, L., Sun, Z., Liu, Q., Dwivedi-Yu, J.,
Yang, Y., Callan, J., and Neubig, G.
\newblock Active retrieval augmented generation.
\newblock In \textit{Proceedings of EMNLP}, 2023.

\bibitem[Jiang et al.(2024)]{jiang2024longllmlingua}
Jiang, H., Wu, Q., Luo, X., Li, D., Lin, C.-Y., Yang, Y., and Qiu, L.
\newblock {LongLLMLingua}: Accelerating and enhancing {LLMs} in long context
  scenarios via prompt compression.
\newblock In \textit{Proceedings of ACL 2024}, pp.\ 1658--1677, 2024.

\bibitem[Liu et al.(2024)]{liu2024kivi}
Liu, Z., Yuan, J., Jin, H., Zhong, S., Xu, Z., Braverman, V.,
Chen, B., and Hu, X.
\newblock {KIVI}: A tuning-free asymmetric 2-bit quantization for {KV} cache.
\newblock In \textit{Proceedings of ICML}, pp.\ 32332--32344, 2024.

\bibitem[Kwon et al.(2023)]{kwon2023pagedattention}
Kwon, W., Li, Z., Zhuang, S., Sheng, Y., Zheng, L., Yu, C.~H.,
Gonzalez, J.~E., Zhang, H., and Stoica, I.
\newblock Efficient memory management for large language model serving with
  {PagedAttention}.
\newblock In \textit{Proceedings of SOSP}, pp.\ 611--626, 2023.

\bibitem[Lewis et al.(2020)]{lewis2020rag}
Lewis, P., Perez, E., Piktus, A., Petroni, F., Karpukhin, V., Goyal, N.,
K\"{u}ttler, H., Lewis, M., Yih, W., Rockt\"{a}schel, T., Riedel, S.,
and Kiela, D.
\newblock Retrieval-augmented generation for knowledge-intensive {NLP} tasks.
\newblock In \textit{Advances in NeurIPS}, vol.\ 33, pp.\ 9459--9474, 2020.

\bibitem[Li et al.(2024a)]{li2024selfroute}
Li, Z., Li, C., Zhang, M., Mei, Q., and Bendersky, M.
\newblock Retrieval augmented generation or long-context {LLMs}?
  {A} comprehensive study and hybrid approach.
\newblock In \textit{Proceedings of EMNLP} (Industry Track), 2024.
\newblock \textit{arXiv preprint arXiv:2407.16833}.

\bibitem[Lin et al.(2024)]{lin2024awq}
Lin, J., Tang, J., Tang, H., Yang, S., Chen, W.-M., Wang, W.-C., Xiao, G.,
Dang, X., Gan, C., and Han, S.
\newblock {AWQ}: Activation-aware weight quantization for {LLM} compression
  and acceleration.
\newblock In \textit{Proceedings of MLSys}, vol.\ 6, 2024.

\bibitem[Ong et al.(2024)]{ong2024routellm}
Ong, I., Almahairi, A., Wu, V., Chiang, W.-L., Wu, T., Gonzalez, J.~E.,
Kadous, M.~W., and Stoica, I.
\newblock {RouteLLM}: Learning to route {LLMs} with preference data.
\newblock \textit{arXiv preprint arXiv:2406.18665}, 2024.

\bibitem[Pan et al.(2024)]{pan2024llmlingua2}
Pan, Z., Wu, Q., Jiang, H., Xia, M., Luo, X., Zhang, J., Lin, Q.,
R\"{u}hle, V., Yang, Y., Lin, C.-Y., Zhao, H.~V., Qiu, L., and Zhang, D.
\newblock {LLMLingua-2}: Data distillation for efficient and faithful
  task-agnostic prompt compression.
\newblock In \textit{Findings of ACL 2024}, pp.\ 963--981, 2024.

\bibitem[Li et al.(2024b)]{li2024snapkv}
Li, Y., Huang, Y., Yang, B., Venkitesh, B., Locatelli, A., Ye, H.,
Cai, T., Lewis, P., and Chen, D.
\newblock {SnapKV}: LLM knows what you are looking for before generation.
\newblock In \textit{Advances in NeurIPS}, vol.~37, 2024.

\bibitem[Robertson \& Zaragoza(2009)]{robertson2009bm25}
Robertson, S. and Zaragoza, H.
\newblock The probabilistic relevance framework: {BM25} and beyond.
\newblock \textit{Foundations and Trends in Information Retrieval},
  3(4):333--389, 2009.

\bibitem[Sheng et al.(2023)]{sheng2023flexgen}
Sheng, Y., Zheng, L., Yuan, B., Li, Z., Ryabinin, M., Chen, B., Liang, P.,
R\'{e}, C., Stoica, I., and Zhang, C.
\newblock {FlexGen}: High-throughput generative inference of large language
  models with a single {GPU}.
\newblock In \textit{Proceedings of ICML}, pp.\ 31094--31116, 2023.

\bibitem[Sparck Jones(1972)]{sparck1972statistical}
Sparck Jones, K.
\newblock A statistical interpretation of term specificity and its application
  in retrieval.
\newblock \textit{Journal of Documentation}, 28(1):11--21, 1972.

\bibitem[Templin(1957)]{templin1957}
Templin, M.~C.
\newblock \textit{Certain Language Skills in Children}.
\newblock Institute of Child Welfare Monograph Series, No.~26,
  University of Minnesota Press, Minneapolis, MN, 1957.

\bibitem[Xiao et al.(2024)]{xiao2024streamingllm}
Xiao, G., Tian, Y., Chen, B., Han, S., and Lewis, M.
\newblock Efficient streaming language models with attention sinks.
\newblock In \textit{Proceedings of ICLR}, 2024.

\bibitem[Xu et al.(2024a)]{xu2023recomp}
Xu, F., Shi, W., and Choi, E.
\newblock {RECOMP}: Improving retrieval-augmented language models with context
  compression.
\newblock In \textit{Proceedings of ICLR}, 2024.

\bibitem[Yu et al.(2024)]{yu2024defenserag}
Yu, T., Xu, A., and Akkiraju, R.
\newblock In defense of {RAG} in the era of long-context language models.
\newblock In \textit{Proceedings of EMNLP}, 2024.

\bibitem[Zhang et al.(2023)]{zhang2023h2o}
Zhang, Z., Sheng, Y., Zhou, T., Chen, T., Zheng, L., Cai, R., Song, Z.,
Tian, Y., R\'{e}, C., Barrett, C., Wang, Z., and Chen, B.
\newblock {H2O}: Heavy-hitter oracle for efficient generative inference of
  large language models.
\newblock In \textit{Advances in NeurIPS}, vol.~36, 2023.

\end{thebibliography}
\end{document}